\documentclass[sigconf, nonacm]{acmart}
\AtBeginDocument{%
	}

\copyrightyear{2023}
\acmYear{2023}
\setcopyright{acmlicensed}\acmConference[MM '23]{Proceedings of the 31st
	ACM International Conference on Multimedia}{October 29-November 3,
	2023}{Ottawa, ON, Canada}
\acmBooktitle{Proceedings of the 31st ACM International Conference on
	Multimedia (MM '23), October 29-November 3, 2023, Ottawa, ON, Canada}
\acmPrice{15.00}
\acmDOI{10.1145/3581783.3612036}
\acmISBN{979-8-4007-0108-5/23/10}

\usepackage{graphicx}
\usepackage{float}
\usepackage{multirow}
\usepackage{array}
\newcolumntype{P}[1]{>{\centering\arraybackslash}p{#1}}
\usepackage{booktabs}
\usepackage[flushleft]{threeparttable}
\usepackage{amsmath}
\usepackage{empheq}
\usepackage[font=normal,skip=2pt]{caption}

\usepackage{balance} 
\begin{document}
	
	\title[FCBoost-Net]{FCBoost-Net: A Generative Network for Synthesizing Multiple Collocated Outfits via Fashion Compatibility Boosting}
	
	\author{Dongliang Zhou}
	\email{zhou-dongliang@outlook.com}
	\orcid{0000-0003-0361-8597}
	\affiliation{%
		\institution{Harbin Institute of Technology, Shenzhen}
		\city{Shenzhen}
		\state{Guandong}
		\country{China}
		\postcode{518055}
	}
	
	\author{Haijun Zhang}
	\orcid{0000-0002-1648-0227}
	\authornote{Corresponding author: Haijun Zhang}
	\email{hjzhang@hit.edu.cn}
	\affiliation{%
		\institution{Harbin Institute of Technology, Shenzhen}
		\city{Shenzhen}
		\state{Guandong}
		\country{China}
		\postcode{518055}
	}
	\author{Jianghong Ma}
	\email{majianghong@hit.edu.cn}
	\orcid{0000-0002-0524-3584}
	\affiliation{%
		\institution{Harbin Institute of Technology, Shenzhen}
		\city{Shenzhen}
		\state{Guandong}
		\country{China}
		\postcode{518055}
	}
	
	\author{Jicong Fan}
	\email{fanjicong@cuhk.edu.cn}
	\orcid{0000-0001-9665-0355}
	\affiliation{%
		\institution{The Chinese University of Hong Kong, Shenzhen}
		\institution{Shenzhen Research Institute of Big Data}
		\city{Shenzhen}
		\state{Guangdong}
		\country{China}
		\postcode{518055}
	}

	\author{Zhao Zhang}
	\email{cszzhang@gmail.com}
	\orcid{0000-0002-5703-7969}
	\affiliation{%
		\institution{Hefei University of Technology}
		\city{Hefei}
		\state{Anhui}
		\country{China}
		\postcode{230009}
	}
	
	\renewcommand{\shortauthors}{Dongliang Zhou, Haijun Zhang, Jianghong Ma, Jicong Fan, \& Zhao Zhang}
	
	\begin{abstract}
		Outfit generation is a challenging task in the field of fashion technology, in which the aim is to create a collocated set of fashion items that complement a given set of items. Previous studies in this area have been limited to generating a unique set of fashion items based on a given set of items, without providing additional options to users. This lack of a diverse range of choices necessitates the development of a more versatile framework. However, when the task of generating collocated and diversified outfits is approached with multimodal image-to-image translation methods, it poses a challenging problem in terms of non-aligned image translation, which is hard to address with existing methods. In this research, we present \textit{FCBoost-Net}, a new framework for outfit generation that leverages the power of pre-trained generative models to produce multiple collocated and diversified outfits. Initially, FCBoost-Net randomly synthesizes multiple sets of fashion items, and the compatibility of the synthesized sets is then improved in several rounds using a novel fashion compatibility booster. This approach was inspired by \textit{boosting} algorithms and allows the performance to be gradually improved in multiple steps. Empirical evidence indicates that the proposed strategy can improve the fashion compatibility of randomly synthesized fashion items as well as maintain their diversity. Extensive experiments confirm the effectiveness of our proposed framework with respect to visual authenticity, diversity, and fashion compatibility.
	\end{abstract}



	
	\maketitle

	\begin{figure}[t]
		\centering
		\includegraphics[width=0.5\textwidth]{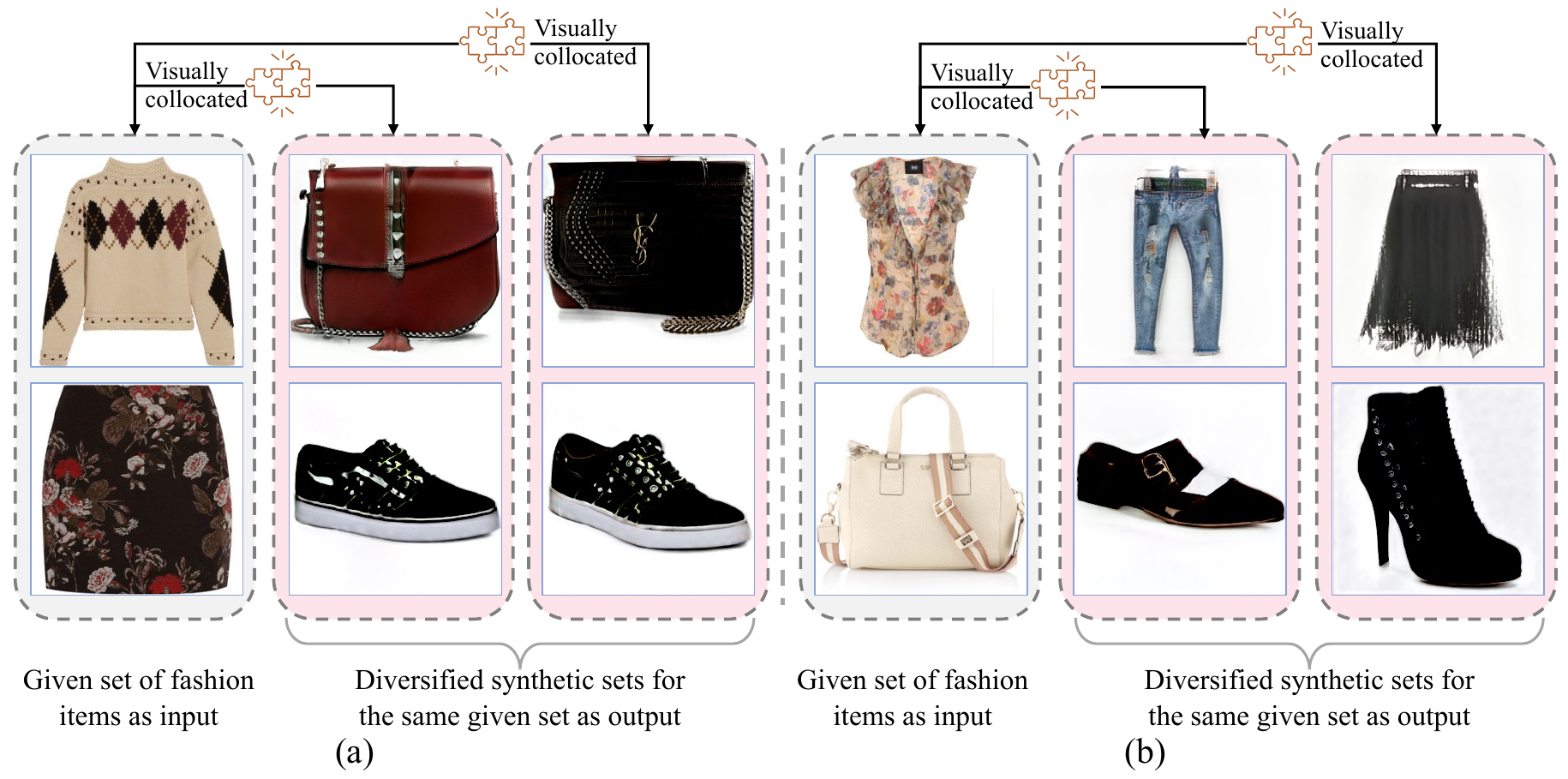}
		\caption{FCBoost-Net is a novel framework that generates multiple diversified sets of fashion items based on a given set of items. Each synthetic set is collocated and can form a complete outfit alongside the given set. Two examples are given here: (a) synthesizing diverse items of bag and shoe that match the given upper and lower clothing; and (b) synthesizing diverse items of lower clothing and shoe that complement the given upper clothing and bag.}
		\label{fig:cover}
	\end{figure}
	
	\section{Introduction}
	
	John Galliano famously proclaimed, ``The joy of dressing is an art." Indeed, the ability to create collocated outfits to achieve specific styles is a coveted skill that many individuals strive to master. Recent advances in generative models, and particularly in regard to generative adversarial networks (GANs) \cite{karras2019style,karras2020analyzing,zhang2020warpclothingout,liu2021clothing,liu2023toward} and diffusion models \cite{ho2020denoising,dhariwal2021diffusion,rombach2022high}, have revolutionized the fashion industry by enabling the synthesis of photo-realistic content with unprecedented quality. These advancements have resulted in a proliferation of applications in the fashion sector, such as virtual try-on \cite{han2018viton,wang2018toward}, draft-to-attire design \cite{han2020design}, and fashion style transfer \cite{yan2022toward}. These developments can enable ordinary individuals to design and create their own outfits with ease. In response to these developments, outfit generation \cite{liu2019toward,liu2019collocating,yu2019personalized,zhou2022learning,zhou2022coutfitgan} has become an increasingly captivating and promising field in recent years. 
	Existing studies of outfit generation can be divided into pair-wise \cite{liu2019toward,liu2019collocating,yu2019personalized} and set-wise \cite{zhou2022learning,zhou2022coutfitgan} approaches. Pair-wise methods focus on generating a collocated item of lower (or upper) clothing from an item of upper (or lower) clothing, whereas set-wise methods concentrate on generating multiple fashion items from different categories, such as upper clothing, bags, lower clothing, and shoes. Set-wise outfit generation is a more generalized approach, as it involves taking a set of fashion items as input and synthesizing a complementary set of items to create a complete outfit. However, existing models, such as OutfitGAN \cite{zhou2022learning} and COutfitGAN \cite{zhou2022coutfitgan}, only offer unique outfit generation and cannot produce diverse fashion items for users. In real fashion design scenarios, users often require multiple choices. To expand the range of collocated fashion items, we propose a new approach to outfit generation that focuses on synthesizing diverse outfits. Based on this, we put forward a framework for outfit generation that is capable of producing multiple distinct sets for a given set of fashion items, with each set forming a visually cohesive outfit when combined with the original set, as illustrated in Fig. \ref{fig:cover}. 
	
	In fact, outfit generation can be approached as a direct multimodal image-to-image (I2I) translation \cite{huang2018munit,DRIT,DRIT_plus,mao2022continuous} task, in which the image data containing extant fashion items are taken as input and the model produces a diverse range of image data representing compatible items. The aim of multimodal I2I translation \cite{huang2018munit,DRIT,DRIT_plus,mao2022continuous} is to transform an image from the source domain to the target domain to generate multiple possibilities. For general multimodal I2I translation, Huang \textit{et al.} \cite{huang2018munit} proposed a multimodal unsupervised I2I translation (MUNIT) framework, which disentangles an input image into content and style codes, and ensures that the style codes fit a Gaussian distribution. In the same period, Lee \textit{et al.} \cite{DRIT} established a framework for diverse I2I translation via disentangled representations (DRIT) to synthesize diverse images in the target domain based on a given image from the source domain. In a later study, they extended their framework into multiple domains \cite{DRIT_plus}. A continuous and diverse I2I translation framework called SAVI2I \cite{mao2022continuous}, is another recent translation framework that can synthesize multiple styles of an image from the source domain to the target domain, with the ability to interpolate the translation. 
	These methods described above adopt a convolutional neural network (CNN)-based encoder-decoder \cite{badrinarayanan2017segnet}  architecture, and rely on  the default assumption that the input and output have an explicit spatial alignment. However, multimodal I2I translation methods are not well-adapted to our outfit generation task, as  they suffer from the following issues: (i) the task of outfit generation necessitates the use of a set of images as input rather than just one, with the output being multiple sets of images rather than individual images, and the number of the images in the input set may be arbitrary; (ii) unlike the default setting of spatial alignment between the input and output in multimodal I2I models (e.g., `summer $\rightarrow$ winter'  and `cat $\rightarrow$ dog' translations \cite{huang2018munit, DRIT, DRIT_plus}), there is no explicit spatial alignment between any two fashion items in an outfit, e.g., upper and lower clothing; and (iii) extant multimodal I2I methods usually apply a disentanglement strategy to decouple images into two different latent codes in an unsupervised manner. However, the task of outfit generation involves a vast number of compatible outfits to provide explicit supervision in terms of fashion compatibility. To address the first issue, there is a need to establish an outfit generation-orientated framework, that is more versatile, and which can handle a set as input and multiple sets as output. In regard to the second issue, many fashion-related translation methods rely on a guided I2I translation that takes clothing silhouette masks \cite{zhou2022learning,zhou2022coutfitgan}, segmentation maps \cite{neuberger2020image}, or human pose landmarks \cite{zheng2019virtually} to overcome the spatial non-alignment between the input and the output. However, the acquisition and application of such additional information can be challenging and may impose an additional burden on users of outfit generation models. To handle the third issue, we need to fully utilize the available data on collocated outfits to build a supervision module to guide the outfit generation process. It is also worth noting that a strong supervision module may limit the diversity of the fashion items generated based on a given set of items. 
	
	In view of the abovementioned issues, we develop a novel framework called \textit{FCBoost-Net}, for the task of outfit generation in which fashion compatibility is enhanced through a boosting strategy. In particular, to reduce the spatial non-alignment between each pair of fashion items in an outfit, we adopt the GAN inversion \cite{zhu2020domain,richardson2021encoding,tov2021designing} technique, which encodes images into the latent space of a pre-trained GAN model. By exploiting the power of a pre-trained GAN model, the training complexity of our framework can be reduced, as we only need to train an encoding network to encode images into a latent code, meaning that image generation is controlled globally without preserving any spatial information. Meanwhile, the theory of local minima \cite{petzka2021non} in deep neural network training suggests that different parameters obtained through training can lead to suboptimal points. Building upon this theory, we believe that different randomly synthesized outfits can be optimized in multiple steps to achieve collocated ensembles. In our approach, a randomly composed set of fashion items can be optimized to be more compatible while maintaining their diversity. To generate outfits, we first synthesize complementary fashion items randomly and then apply a fashion compatibility boosting strategy to enable our framework to consistently improve the compatibility of the generated outfits while ensuring that the diversity is preserved. In particular, our strategy is inspired by \textit{boosting} algorithms \cite{kearns1994cryptographic,chen2016xgboost}, which can progressively enhance the performance of an algorithm through a gradual process with multiple steps. In particular, their fashion compatibility is improved in the recurrent steps through the application of the proposed fashion compatibility booster. Motivated by the Bellman–Ford algorithm \cite{bellman1958routing}, set-wise generation of outfits can be viewed as a graph in which each fashion item represents a node. A ``relaxation'' operation is applied to the graph to find a shorter path than before, in an iterative manner. The distance between each pair of fashion items in the same outfit can be relaxed to obtain better fashion compatibility. To achieve this goal, we design a fashion compatibility booster module, which compares the distance between the current pair of fashion items and the previously generated pair in the fashion compatibility space. In addition, to maintain the diversity of results, different synthetic sets of fashion items based on the same given set of items are pushed away with a diversity loss. After training, each randomly synthesized fashion item becomes more compatible with the existing items than before with different optimums.
	As shown in Fig. \ref{fig:cover},  our framework can synthesize multiple diversified sets of fashion items that match the given set of items. To further validate our model, extensive experiments were conducted on the OutfitSet dataset \cite{zhou2022coutfitgan}, and the results demonstrated the effectiveness of our method. The main contributions of this research can be summarized as follows:  
	\begin{enumerate}
		\item We raise a new problem of collocated and diversified outfit generation. To the best of our knowledge, we are the first to address this problem through the use of a boosting framework. Our framework leverages a GAN inversion technique to overcome the issue of spatial non-alignment between fashion items in an outfit.
		\item We develop a fashion compatibility booster module to enhance the compatibility of different synthesized outfits in a recurrent manner while simultaneously preserving their diversity through the implementation of a diversity loss.
		\item Extensive experiments demonstrated that our framework outperforms other multimodal I2I translation methods in terms of visual authenticity, diversity, and fashion compatibility. This suggests that our model can be used as a valuable tool for diversified outfit generation by fashion designers. 
	\end{enumerate}
	
	
	\section{Related Work}
	\label{sec:bg}
	
	Outfit generation is a multifaceted research area that has been investigated in several studies. In this section, we present a brief review of previous work on outfit generation and multimodal I2I translation. Finally, we highlight the positioning of our work. 
	
	\textbf{Outfit Generation:} In general, outfit generation attempts to synthesize complementary fashion items that match given fashion items. In previous studies, pair-wise \cite{liu2019toward,liu2019collocating,yu2019personalized} and set-wise \cite{zhou2022learning,zhou2022coutfitgan} outfit generation methods have been explored. For instance, an Attribute-GAN \cite{liu2019toward} framework was proposed to carry out the outfit generation in a pair-wise way with a new attribute loss. This model adopted upper and lower clothing as the domains of translation. Liu \textit{et al.} \cite{liu2019collocating} improved Attribute-GAN by introducing a multi-discriminator framework to further enhance the quality of collocated clothing generation. In contrast, Yu \textit{et al.} \cite{yu2019personalized} proposed a new task in the field of outfit generation, personalized fashion design, in which users were recommended collocated clothing synthesized by their model. These generative models only worked on upper and lower clothing domains, meaning that they needed to learn a mapping only between these two domains. To overcome this issue, Zhou \textit{et al.} \cite{zhou2022learning} originally proposed a new framework called OutfitGAN to learn a one-to-many mapping for outfit generation. OutfitGAN adopted one fashion item as input and produced multiple items from different categories as output. Zhou \textit{et al.} \cite{zhou2022coutfitgan} then extended their framework to a many-to-many mapping that was capable of synthesizing complementary fashion items based on an arbitrary set of items as input. These set-wise frameworks concentrated on the generation of outfits consisting of several fashion items.
	
	\textbf{Multimodal I2I Translation:} In this approach, a model learns a conditional distribution that can produce diverse images in a target domain, given an input image in the source domain as input. Huang \textit{et al.} \cite{huang2018munit} proposed a model called MUNIT to disentangle images into style and content codes. Once a well-trained disentanglement network had been achieved, content code from the source domain could be aggregated with the style code sampled from a Gaussian distribution and decoded into an image in the target domain. In the same period, Lee \textit{et al.} \cite{DRIT} proposed a scheme called DRIT to disentangle images into attribute and content space. In this approach, the source and target domains shared the same content domain. In a later study, Lee \textit{et al.} \cite{DRIT_plus} extended DRIT to create the DRIT++ framework, which facilitated translation across multiple domains. More recently, Mao \textit{et al.} \cite{mao2022continuous} utilized signed attribute vectors to translate an image from the source domain to the target domain with diverse possibilities alongside continuous translation through interpolation.
	
	\textbf{Positioning of Our Work:} The aim of our research is to synthesize collocated and diversified sets of fashion items based on a given set of items, which constitutes a novel task in the field of fashion intelligence. COutfitGAN \cite{zhou2022coutfitgan} is the most relevant existing scheme in this context; however, the lack of diversity of the generated outfits and the need for additional guiding information in the form of silhouette masks during the translation process imposes certain limitations on this approach. Multimodal I2I translation methods \cite{huang2018munit, DRIT, DRIT_plus, mao2022continuous} could be explored as an alternative approach to outfit generation, but they are not directly applicable to our task due to their dependence on an assumption of spatial alignment between the input and output. Furthermore, these methods can only use unsupervised learning to create mappings. In contrast, our task provides explicit supervision by using outfits put together by fashion experts.
	
	\section{FCBoost-Net}
	\label{sec:method}
	In this section, we first present a formulation of the problem considered here. We then describe the key components of our framework and set out the training objectives for our model.
	
	\subsection{Problem Formulation}
	
	In practice, fashion designers may meet the demands of their clientele by producing multiple fashion items that complement their preselected choices, and may offer a broader range of options according to their different dressing styles. For instance, a white T-shirt could be paired with light blue jeans or dark straight-leg pants, resulting in a casual or business fashion style, respectively. It is crucial to take into account the possible preferences of their customers when providing them with choices to supplement their current wardrobe. To formulate this process, we can define a universal set of fashion items as $\mathcal{O}=\{ \mathcal{O}_i \}_{i=1}^{N}$, including $N$ fashion items of different categories, forms a complete outfit. Upper clothing, bags, lower clothing, and shoes are the four categories of fashion items that are predefined in this paper. For any possible given set $\{\mathcal{O}_i\}_{i=N_1}^{N_g}$ including $N_g$ fashion item sampled from $\mathcal{O}$, where $0 < N_g < N$, our outfit generation framework aims to synthesize $K$ possible complementary sets. To ensure that our framework can synthesize diverse sets of fashion items, random noise is fed into our model in a similar way to many in previous multimodal I2I translation methods \cite{huang2018munit,DRIT,DRIT_plus,mao2022continuous}. Formally, given a set of $K$ latent codes $\mathbf{z}_1,\cdots,\mathbf{z}_K\in\mathcal{Z}$, where $\mathcal{Z}$ is a Gaussian distribution, and a given set of $N_g$ fashion items $\{\mathcal{O}_i\}_{i=N_1}^{N_g}$, the mapping function is represented as $F:(\{  \mathcal{O}_{N_1},\cdots, \mathcal{O}_{N_g} \},\mathbf{z}_k ) \mapsto \{\widetilde{\mathcal{O}}_i\}_{i=N_{k,1}}^{N_{k,s}}$, where $ k \in \{1,\cdots,K\}$ and $\{N_1,\cdots, N_g,N_{k,1},\cdots,N_{k,s}\}=\{i\}_{i=1}^{N}$. Since we aim to develop an accurate model for synthesizing  collocated and diversified outfits based on a given set of multiple fashion items, there are three goals that we need to achieve: (i) each item in a synthetic set is expected to be visually plausible, which means $\widetilde{\mathcal{O}}_{i}$ should fit the distribution of ${\mathcal{O}_i}$; (ii) the synthetic sets of fashion items are expected to be diversified from the perspective of human observation, which means that there is a large difference between $ \widetilde{\mathcal{O}}_{i}^{k_1}$ and $ \widetilde{\mathcal{O}}_{i}^{k_2}$  in terms of human visual perspective if there is a large distance between $\mathbf{z}_{k_1}$ and $\mathbf{z}_{k_2}$; and (iii) each synthetic set of fashion items should be compatible with the given set of items, meaning that $\{ \mathcal{O}_{N_1},\cdots, \mathcal{O}_{N_g},\widetilde{\mathcal{O}}_{N_{k,1}},\cdots,\widetilde{\mathcal{O}}_{N_{k,s}}\}$ is expected to represent a visually collocated outfit.

	\subsection{Proposed Framework}
	
	\begin{figure}[t]
		\centering
		\includegraphics[width=0.5\textwidth]{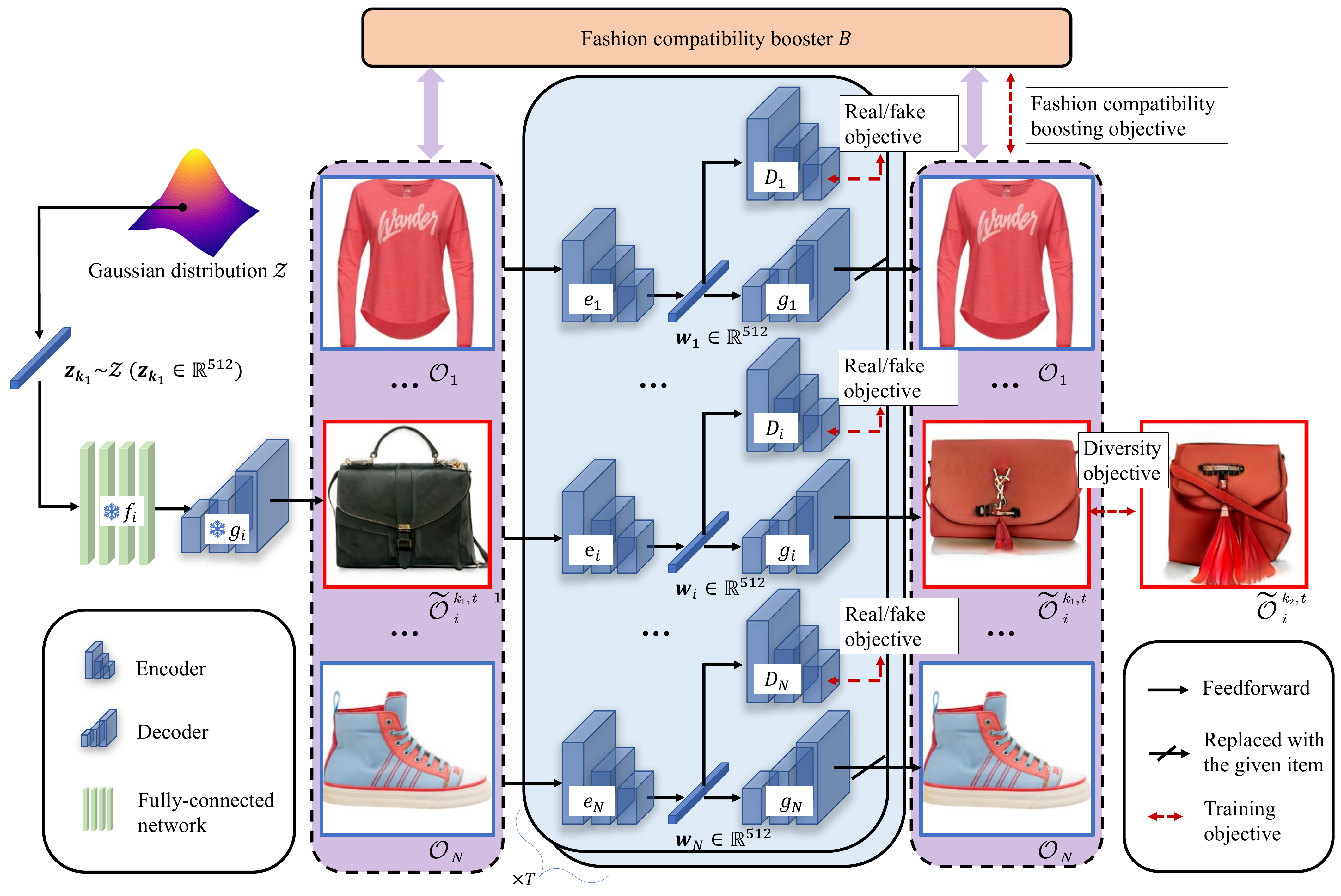}
		\caption{The overall pipeline of FCBoost-Net. (Here, the fashion items labeled with blue boxes represent the given items, while those enclosed within red boxes indicate the synthesized items.)}
		\label{fig:overview}
	
	\end{figure}
	
	In this section, we first describe the overall pipeline and key components for our proposed framework, and then describe its key components in more detail.
	
	\textbf{Overall Pipeline.} Prior methods of multimodal I2I translation \cite{huang2018munit,DRIT,DRIT_plus,mao2022continuous} have encountered challenges in terms of addressing semantic non-alignment issues due to the use of encoder-decoder architectures. To overcome these difficulties, we propose a new generation pipeline called FCBoost-Net, which is built upon a pre-trained StyleGAN \cite{karras2019style,karras2020analyzing}. To ensure that this paper is self-contained, we give a brief introduction to StyleGAN here. The generator of StyleGAN consists of two key components: a mapping network $f$ and a synthesis network $g$. The mapping network $f$ is responsible for mapping a noise vector $\mathbf{z}$ sampled from a Gaussian distribution $\mathcal{Z}$ to a disentangled space $\mathcal{W}$, as $\mathbf{w} \in \mathbb{R}^{512}$. Then, $\mathbf{w}$ is used to progressively control the synthesis network $g$ to generate an image $\mathbf{x}\in \mathbb{R}^{H\times W \times 3}$. Recent studies have shown that StyleGAN inversion \cite{zhu2020domain,richardson2021encoding,tov2021designing} is a valuable technique for image-to-image translation, especially for human face editing, as it employs a new encoding network in addition to the pre-trained StyleGAN to encode the image into $\mathcal{W}$ space (although some methods \cite{richardson2021encoding,tov2021designing} have adopted an extended $\mathcal{W}+$ space). These methods apply manipulation to these encoded latent codes to obtain edited latent codes which are fed into the synthesis network $g$ to generate edited images. It is worth noting that GAN inversion techniques can only be applied in the same domain. During the training phase, the input and output must be kept as similar as  possible to the training objectives, using loss functions such as L1 \cite{isola2017image} or learned perceptual image patch similarity (LPIPS) \cite{zhang2018unreasonable}. To create a complete outfit that includes $N$ fashion items from different categories, our framework needs to synthesize multiple complementary sets of items, based on a given set of items that forms a subset of the complete outfit. As shown in Fig. \ref{fig:overview}, our proposed framework consists of $N$ separate generators, discriminators, and a fashion compatibility booster. To obtain pre-trained modules of generators, mapping and synthesis networks were trained on the training dataset. For the $i$-th category fashion item, we pre-trained a mapping network $f_i$ and a synthesis network $g_i$ in an unconditional generation manner \footnote{The implementation of the pre-training stage was based on \url{https://github.com/NVlabs/stylegan2-ada-pytorch}}. After pre-training the $i$-th mapping network $f_i$ and the $i$-th synthesis network $g_i$, our framework can synthesize an arbitrary $i$-th fashion item for an outfit and takes a sample noise $\mathbf{z}_k \sim \mathcal{Z}$ as input, i.e., $f_i \circ g_i (\mathbf{z}_k) \mapsto \widetilde{\mathcal{O}}_i ^ {k,0}$. However, at this stage, the outfit does not necessarily fit together visually. To further improve the fashion compatibility of the current outfit, we encode it as a latent code and decode it into a new outfit through a supervised process, with a pre-trained fashion compatibility booster that is applied recurrently. More specifically, the fashion compatibility booster is designed to compare the distance between the current pair of items and between a single current item with another item from the last round. The feedforward process of FCBoost-Net takes randomly synthesized fashion items and the given items as input, and synthesizes new fashion items recurrently as follows:
	\begin{empheq}[left = \empheqlbrace]{align}
		\begin{split}
			\widetilde{\mathcal{O}}_i ^ {k,0} = ~ & f_i \circ g_i (\mathbf{z}_k), \\ 
			\widetilde{\mathcal{O}}_i ^ {k,t}= ~ & e_i \circ g_i (( \mathcal{O}_{N_1}\oplus \cdots\oplus \mathcal{O}_{N_g} \\
			& \oplus  \widetilde{\mathcal{O}}_{N_{k,1}} ^ {k,t-1}\oplus \cdots \oplus \widetilde{\mathcal{O}}_{N_{k,s}} ^ {k,t-1})), \quad t=1,\cdots,T,
		\end{split}
	\end{empheq}
	where $\oplus$ is a concatenation operation among images in channel dimension, and $\widetilde{\mathcal{O}}_i ^ {k,T}$ is the final synthesized fashion item.
	
	\textbf{Generator.} The generation module $G$ consists of $N$ separate generators. To synthesize the $i$-th fashion item $\widetilde{\mathcal{O}}_i$, each generator $G_i$ uses three key components: a mapping network $f_i$, a synthesis network $g_i$, and an encoding network $e_i$. The generators are tasked with translating a given set of fashion items $\{\mathcal{O}_{N_1}, \cdots, \mathcal{O}_{N_g}\}$ and a set of sampled latent codes $\{\mathbf{z_k}\}_{k=1}^{K}$ into $K$ sets of synthesized fashion items $ \{\widetilde{\mathcal{O}}_{N_{k,1}}, \cdots, \widetilde{\mathcal{O}}_{N_{k,s}}\}_{k=1}^{K} $. The mapping network $f_i$ and synthesis network $g_i$ are pre-trained on the $i$-th fashion item using the training set. The detailed encoding network $e_i$ is adapted from a pixel-to-style-to-pixel (pSp) \cite{richardson2021encoding} network. In our implementation, $e_i$ takes an outfit concatenated in the channel dimension as input and produces a latent code in $\mathbb{R}^{512}$ space. During the training phase, each mapping network $f_i$ is responsible for generating real latent codes as real samples for the discriminator $D_i$, and these latent codes are also fed to the corresponding synthesis network $g_i$ to generate randomly synthesized fashion items in advance. Each encoding network $e_i$ is responsible for ensuring that the current outfit is encoded into a new latent code to improve the fashion compatibility over each subsequent round of fashion compatibility boosting. Each synthesis network $g_i$ is responsible for generating the $i$-th fashion item with latent code from $f_i$ or $e_i$.
	
	\textbf{Discriminator.} To ensure that the encoded latent codes from the encoding networks can be fed to synthesis networks to generate photo-realistic fashion items, our framework consists of $N$ separate discriminators, which are used to supervise the encoding networks. The $i$-th discriminator $D_i$ is responsible for ensuring that the latent codes encoded by $e_i$ fit the latent space $\mathcal{W}_i$ of the pre-trained generator $G_i$. In particular, they distinguish the latent codes generated from mapping networks as real samples and latent codes from encoding networks as fake samples. Each discriminator $D_i$ consists of four fully-connected layers with leaky rectified linear unit (ReLU) activations except for the last layer. To enhance the stability of the training process, a discriminator pool is applied to keep the generated and encoded latent codes from the previous step.
	
	\textbf{Fashion Compatibility Booster.} When an outfit has been composed, the fashion compatibility booster $B$ calculates the distance between each pair of items from a visual compatibility perspective. The fashion compatibility booster is pre-trained in advance. Following the work in \cite{vasileva2018learning}, our fashion compatibility module consists of a feature extractor, which maps images of fashion items into embeddings, and an embedding mapping network which maps an embedding into type-aware embedding space. For each pair of fashion items $(\mathcal{O}_i, \mathcal{O}_j)$, the distance between them in type-aware space reflects their compatibility. The fashion compatibility booster module is pre-trained to supervise the training of our generator. During the pre-training phase of the fashion compatibility booster, we utilized the composed outfits from the training set as positive samples and randomly composed outfits as negative samples. During the training phase for FCBoost-Net, all of the parameters of the fashion compatibility booster were frozen to allow us to compare the distance between a certain fashion item $\widetilde{\mathcal{O}}_i ^ {t}$ in the $t$-th boosting round and another fashion item $\widetilde{\mathcal{O}}_i ^ {t}$ in the same round and the distance of $\widetilde{\mathcal{O}}_i ^ {t}$ with another fashion item $\widetilde{\mathcal{O}}_i ^ {t-1}$ in the $(t-1)$-th boosting round. The relaxation operation in the outfit generation task ensures that the outfits synthesized in the $t$-th round have better fashion compatibility in comparison to those of the $(t-1)$-th round. It should be noted that we introduce a comparison between the current and past status of each outfit to further improve fashion compatibility in a recurrent process. In this manner, randomly synthesized sets of fashion items become more compatible at each step.
	
	\subsection{Training Objectives}
	
	Given an outfit $\mathcal{O}$ including $N$ fashion items $ \{ \mathcal{O}_i \}_{i=1}^{N}$ and a batch of noise vectors $\{\mathbf{z}_k\}_{k=1}^{K}$ sampled from $\mathcal{Z}$ where $K > 1$, we train our framework using the following objectives.
	
	\textbf{Real/Fake Objective.} During training, we randomly sample a batch of latent codes $\{\mathbf{z}_k\}_{k=1}^{K}$ and generate two types of latent codes: $\mathbf{w}_i^{k}=f_i(\mathbf{z}_k)$ from mapping networks, and $e_i ( \mathcal{O}_{N_1}\oplus \cdots\oplus \mathcal{O}_{N_g} 
	\oplus  \widetilde{\mathcal{O}}_{N_{k,1}} ^ {k,t-1}\oplus \cdots \oplus \widetilde{\mathcal{O}}_{N_{k,S}} ^ {k,t-1})$ from encoding networks, where $t\in \{1,\cdots,T\}$. Each discriminator $D_i$ takes these latent codes as input and learns to distinguish between the original latent codes and encoded latent codes with the following training objective:
	\begin{align}
		\begin{split}
			\mathcal{L}_{dis}^{t} =  \frac{1}{N} \sum_{i=1}^{N}&\mathbb{E}_{\mathbf{z}_k\sim \mathcal{Z}, \mathcal{O}_{N_1}, \cdots, \mathcal{O}_{N_g} \sim \mathcal{O}}[\log (D_i(f_i(\mathbf{z}_k)))\\
			&+ \log(1-  D_i(e_i (\widetilde{O}_{i}^{k,t})))] ,
		\end{split}
	\end{align}
	where $\widetilde{O}_{i}^{k,t}$ is the $i$-th fashion item in the synthesized outfit based on the latent code $\mathbf{z}_k$ in the $t$-th boosting round. For the $t$-th round of latent codes from the encoding networks, the discriminators provide an adversarial loss as:
	\begin{equation}
		\begin{array}{ll}
			\mathcal{L}_{adv}^{t} = \frac{1}{N} \sum_{i=1}^{N}\mathbb{E}_{\mathbf{z}_k\sim \mathcal{Z}, \mathcal{O}_{N_1}, \cdots, \mathcal{O}_{N_g} \sim \mathcal{O}}[\log (D_i(e_i (\widetilde{O}_{i}^{k,t})))] .
		\end{array}
	\end{equation}
	
	Using the training loss, our generator can produce $\mathbf{w}_i$, which lies in $\mathcal{W}_i$ space. This real/fake objective is designed to guide our generator to synthesize photo-realistic images of fashion items.
	
	\textbf{Diversity Objective.} To increase the diversity of the different synthesized sets based on a batch of random latent codes $\{\mathbf{z}_k\}_{k=1}^{K}$, we explicitly regularize $G$ with the following diversity objective, following the scheme in \cite{choi2020starganv2}:
	\begin{equation}
		\mathcal{L}_{div}^{t}=-\frac{1}{N} \sum_{i=1}^{N} \mathbb{E}_{\mathbf{z}_{k_1}\neq \mathbf{z}_{k_2}}[\mathbf{d}(\widetilde{O}^{k_1,t}_i, \widetilde{O}^{k_2,t}_i)],
	\end{equation}
	where $\mathbf{d}(\cdot,\cdot)$ is a distance function. The diversity objective encourages $G$ to synthesize different fashion items in the same category based on different latent codes. In our implementation, we adopt LPIPS \cite{zhang2018unreasonable} as our distance metric.
	
	\textbf{Fashion Compatibility Boosting Objective}. FCBoost-Net adopts a boosting process to improve the last synthesized outfit and to create a better outfit with higher fashion compatibility. For the $t$-th round of training $G$, the fashion compatibility boosting objective becomes:
	\begin{align}
		\label{eq:fcb}
		\begin{split}
			\mathcal{L}_{fcb}^{t}=\mathbb{E}_{i,j\in \{1,\cdots,N\},i\neq j, k\in \{1,\cdots,K\}}&\max  \{0, B(\widetilde{\mathcal{O}}_i ^ {k,t}, \widetilde{\mathcal{O}}_j ^ {k,t})-B(\widetilde{\mathcal{O}}_i ^ {k,t},\\
			& sg(\widetilde{\mathcal{O}}_j ^ {k,t-1}))+ \alpha\},
		\end{split}
	\end{align}
	where $sg$ stands for a stopgradient operator that is used to eliminate the chain effects of fashion compatibility loss, and $\alpha$ is some margin. This loss objective is designed to improve the fashion compatibility of the current outfit in an iterative way.
	
	\textbf{Full Objective.} When training the generator $G$ of our FCBoost-Net, the full objective functions can be summarized as follows:
	\begin{equation}
		\label{eq:total_loss}
		\mathcal{L}_{total}= \mathbb{E}_{t\sim \{1,\cdots,T\}} (\mathcal{L}_{adv}^{t} + \lambda_{div}\mathcal{L}_{div}^{t}+\lambda_{fcb}\mathcal{L}_{fcb}^{t}),
	\end{equation}
	where $\mathcal{L}_{adv}^{t}$ is the GAN loss in the $t$-th round, which is used to ensure the visual authenticity of the synthetic fashion items; $\mathcal{L}_{div}^{t}$ is the diversity loss in the $t$-th round, which is used to ensure that the generator $G$ synthesizes diversified results; and $\mathcal{L}_{fcb}^{t}$ is a fashion compatibility boosting loss that is applied to improve the fashion  compatibility of synthetic outfits between rounds $(t-1)$ and $t$. $\lambda_{div}$ and $\lambda_{fcb}$ are hyper-parameters that are used to balance the relative contributions of the loss terms.
	
	\section{Experiments}
	\label{sec:exp}
	
	\subsection{Implementation Details}
	
	\textbf{Dataset.} To examine the performance of our method, we used the OutfitSet dataset \cite{zhou2022coutfitgan}, which contains 20,000 sets of visually collocated fashion item images with the size of $256\times 256$ pixels. Each set consists of a visually collocated outfit chosen by fashion experts from Polyvore.com and shows fashion items in four categories: upper clothing, bags, lower clothing, and shoes. The FCBoost-Net model was trained on a subset of 16,000 (80\%) of these sets, while the remaining 4,000 (20\%) sets were used during the test phase. To fully evaluate the proposed framework, each fashion item in a set was randomly masked to construct three settings, i.e., to be given one, two, or three fashion items.
	
	\textbf{Network Training.}  In our implementation, the batch size for training of FCBoost-Net was set to four. All experiments were performed on a single A6000 graphics card, and the implementation was carried out in PyTorch \cite{paszke2017automatic}.  FCBoost-Net was optimized with the Adam \cite{kingma2014adam} optimizer with $\beta_1=0$ and $\beta_2=0.99$, and its learning rate was set to $2\times 10 ^{-4}$. The number of training iterations for the model was set to $60,000$. The margin $\alpha$ in Eq. (\ref{eq:fcb}) was set to $0.2$. We set the coefficients empirically to balance the losses as follows:  $\lambda_{div}=10$ and $\lambda_{fcb}=20$  in Eq. (\ref{eq:total_loss}). The number of boosting rounds $T$ was set to two in our implementation.
	
	\subsection{Comparisons}
	
	\textbf{Baselines}. Since there are few other models that can carry out the task considered here, we chose the following benchmark methods for comparison with the proposed FCBoost-Net, as these are most closely related: MUNIT \cite{huang2018munit}, DRIT \cite{DRIT}, DRIT++ \cite{DRIT_plus}, and SAVI2I \cite{mao2022continuous}. These baseline methods were implemented by leveraging the original codes made available by the respective authors, and the hyper-parameters were meticulously tuned to the highest degree possible, in order to optimize the outfit generation task.
	
	\textbf{Evaluation Metrics}. We applied a variety of evaluation criteria to compare our proposed FCBoost-Net with the other state-of-the-art models in terms of visual authenticity, diversity, and fashion compatibility. To evaluate the visual authenticity of the synthesized images, we employed the Fr\'{e}chet inception distance (FID) \cite{heusel2017gans}, which has been suggested by previous authors \cite{zhou2022learning,zhou2022coutfitgan,DRIT_plus,mao2022continuous} as a reliable measure for gauging the visual authenticity between real images and synthetic images. A lower FID score means better visual authenticity. To measure the diversity of the synthesized images of fashion items, we applied the LPIPS \cite{zhang2018unreasonable} metric, following the works in \cite{DRIT,DRIT_plus,choi2020starganv2,mao2022continuous}. A higher LPIPS score indicates better diversity. To assess the fashion compatibility of the synthesized images, we utilized `fashion fill-in-the-blank best times' (F$^2$BT) \cite{zhou2022coutfitgan}, which reduces the subjectivity of the evaluation of fashion compatibility and allows us to gauge the degree of compatibility between the given set and synthetic set of fashion items. F$^2$BT is a ranking metric that counts the times of best fashion compatibility among all counterpart methods for each method, using a pre-trained fashion compatibility evaluation model provided by MMFashion \cite{mmfashion}. A higher F$^2$BT score indicates better fashion compatibility.
	
	\begin{figure}[t]
		\centering
		\includegraphics[width=0.5\textwidth]{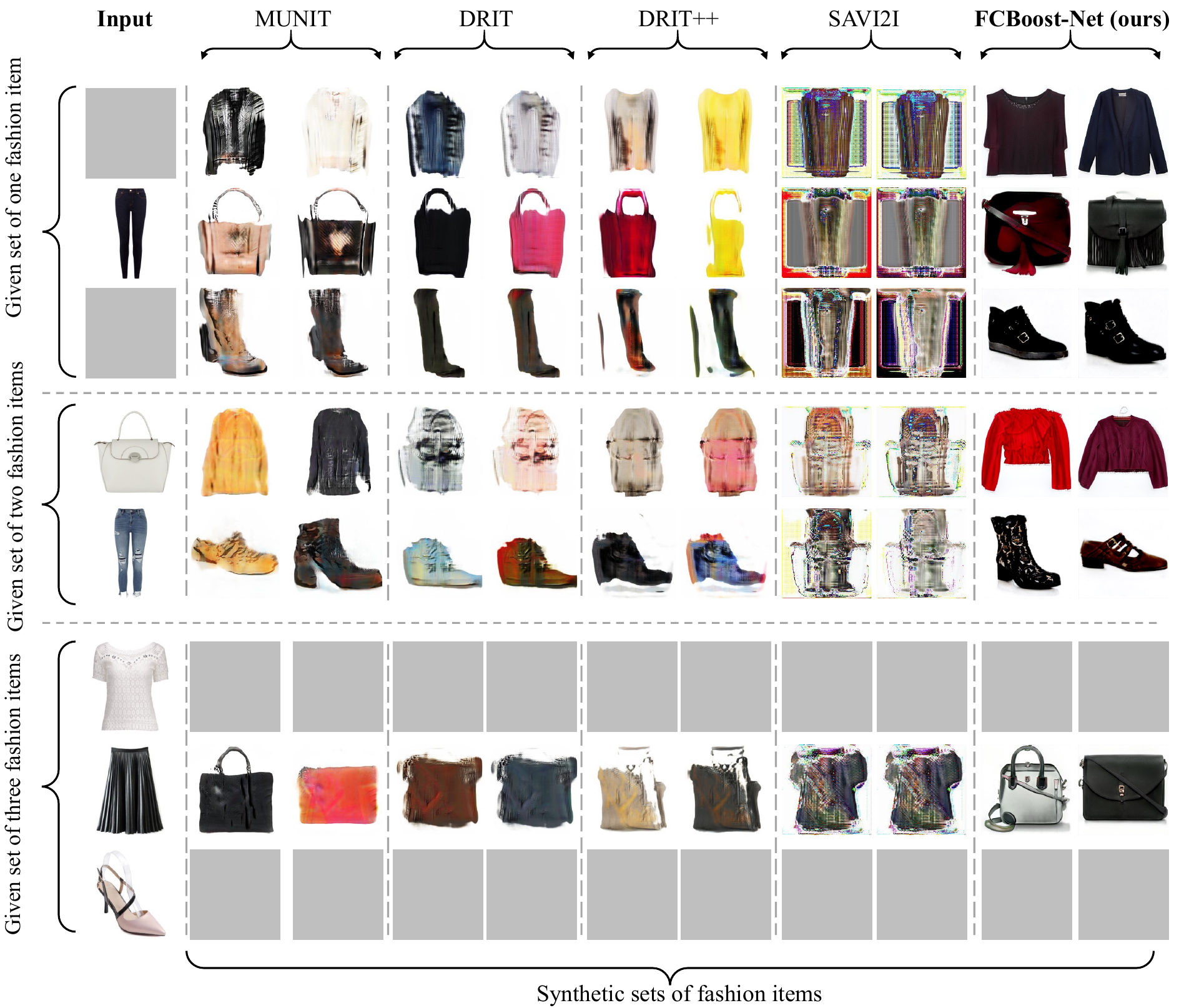}
		\caption{Comparisons between our FCBoost-Net and other multimodal I2I translation methods, which are MUNIT \cite{huang2018munit}, DRIT \cite{DRIT}, DRIT++ \cite{DRIT_plus}, and SAVI2I \cite{mao2022continuous}, for input based on one, two, or three fashion items. (Zoom in for a better view.)}
		\label{fig:cmp_samples}
	\end{figure}
	
	\textbf{Qualitative Comparison.} We present some qualitative comparisons for the task of outfit generation, which is significantly more challenging than general multimodal I2I translation, as it requires the model to learn large variations in spatial nonalignment and appearance. The results of this comparison, based on input consisting of one, two, and three fashion items, are depicted in Fig. \ref{fig:cmp_samples}. The findings show that our model yields the most photo-realistic sets of fashion items. The other baselines, MUNIT, DRIT, and DRIT++, fail to produce photo-realistic results, and the diversity in the items generated by these models is mainly related to the colors of items and rarely the shapes. Another noteworthy aspect is that SAVI2I struggles to produce basic silhouettes for the fashion items, which can be attributed to the fact that it employs linear interpolation in the attribute vector used for continuous translation in the nonalignment task. Fig. \ref{fig:cmp_samples} shows that our model generates multiple sets of fashion items with better visual authenticity, diversity, and fashion compatibility than the alternatives.

	\begin{table}[h]
		\centering
		\small
		\caption{Comparison of FCBoost-Net with baselines in terms of visual authenticity}
		\label{tab:fid}
		\begin{tabular}{p{2.7cm} P{0.88cm} P{0.88cm} P{0.88cm} P{0.88cm} }
			\toprule
			\multirow{2}{*}{\textbf{Method}} & \multicolumn{4}{c}{\textbf{FID}($\downarrow$)} \\
			& 1&2 &3 &Avg.\\
			\hline 
			MUNIT \cite{huang2018munit} & 144.123 & 109.487 & 114.031 & 122.547 \\
			DRIT \cite{DRIT} & 174.499 & 170.483 & 177.081 & 174.021 \\
			DRIT++ \cite{DRIT_plus} & 160.422 & 163.235 & 171.075 & 164.910 \\
			SAVI2I \cite{mao2022continuous} & 259.850 & 260.779 & 278.122 & 266.250 \\
			\hline
			FCBoost-Net (ours) & \textbf{71.102} & \textbf{69.010} & \textbf{69.695} & \textbf{69.935} \\
			\bottomrule
		\end{tabular}
		\begin{tablenotes}
			\small
			\item * Note: The meaning of the values ``1, 2, 3''  pertains to the number of the given fashion items being considered. In addition, the term ``Avg." signifies the average performance across all three scenarios. 
		\end{tablenotes}
	\end{table}
	
	\begin{table}[h]
		\centering
		\small
		\caption{Comparison of FCBoost-Net with baselines in terms of diversity}
		\label{tab:lpips}
		\begin{tabular}{p{2.7cm} P{0.88cm} P{0.88cm} P{0.88cm} P{0.88cm} }
			\toprule
			\multirow{2}{*}{\textbf{Method}} & \multicolumn{4}{c}{\textbf{LPIPS}($\uparrow$)} \\
			& 1&2 &3 &Avg.\\
			\hline 
			MUNIT \cite{huang2018munit} & 0.431 & 0.431 & 0.423 &0.428  \\
			DRIT \cite{DRIT} & 0.263 &  0.234 &  0.223 & 0.240 \\
			DRIT++ \cite{DRIT_plus} & 0.284 & 0.293 & 0.296 &  0.291 \\
			SAVI2I \cite{mao2022continuous} & 0.286 & 0.222 & 0.152 & 0.220 \\
			\hline
			FCBoost-Net (ours) & \textbf{0.521} & \textbf{0.499} & \textbf{0.454} & \textbf{0.491} \\
			\bottomrule
		\end{tabular}
	\end{table}

	\begin{table}[h]
		\centering
		\small
		\caption{Comparison of FCBoost-Net with baselines in terms of fashion compatibility}
		\label{tab:f2bt}
		\begin{tabular}{p{2.7cm} P{0.88cm} P{0.88cm} P{0.88cm} P{0.88cm} }
			\toprule
			\multirow{2}{*}{\textbf{Method}} & \multicolumn{4}{c}{\textbf{F$^2$BT}($\uparrow$)} \\
			& 1&2 &3 &Avg.\\
			\hline 
			MUNIT \cite{huang2018munit} & 19.9  \% & 21.7\% & 21.0\% & 20.7\%  \\
			DRIT \cite{DRIT} & 12.5 \% & 16.1\% & 20.8\% & 16.5\% \\
			DRIT++ \cite{DRIT_plus} & 14.4 \% & 17.1\% & 17.8\% & 16.4\% \\
			SAVI2I \cite{mao2022continuous} & 20.1 \% & 14.5\% & 11.5\% & 15.4\% \\
			\hline
			FCBoost-Net (ours) & \textbf{33.1\%} & \textbf{30.6\%} & \textbf{29.3\%} & \textbf{31.0\%} \\
			\bottomrule
		\end{tabular}
		\begin{tablenotes}
			\small
			\item * Note: Percentages may not sum to 100 due to the use of rounding.
		\end{tablenotes}
	\end{table}
	\textbf{Quantitative Comparison.} We also carried out a quantitative analysis of outfit generation, with a focus on visual authenticity, diversity, and fashion compatibility, as shown in Tables \ref{tab:fid}, \ref{tab:lpips}, and \ref{tab:f2bt}, respectively. The results indicate that our model outperforms the other baselines by a large margin across all three metrics. In terms of visual authenticity, our method has the best translation ability, and produces highly photo-realistic results, as evidenced by the superior FID metric score shown in Table  \ref{tab:fid}. This can be attributed to the use of a vector-based latent code that enables FCBoost-Net to generate compatible fashion items globally, bypassing the constraints of an encoder-decoder architecture, which hinders multimodal I2I translation methods without spatial alignment between the fashion pieces in an outfit. From the results for the diversity metric, we see that our model achieves higher diversity for the generated outfits compared to the other baselines, as demonstrated by the superior LPIPS metric score in Table \ref{tab:lpips}. Finally, regarding fashion compatibility, our model generates a more compatible and visually appealing combination of fashion items compared to the other baselines, as indicated by the better F$^2$BT metric score in Table \ref{tab:f2bt}. We found that FCBoost-Net benefits from the fashion compatibility booster module, which significantly enhances the compatibility of the synthesized outfits compared to the other baselines.

	\subsection{Ablation Study}
	
	To analyze each component of FCBoost-Net, we conducted ablation studies of the effectiveness of the diversity loss, the fashion compatibility booster, and the boosting process.
	\begin{table}[t]
		\centering
		\small
		\caption{Comparison of the diversity in the results from FCBoost-Net with those from a variant without the diversity loss function (LPIPS)}
		\label{tab:lpips_abla}
		\begin{tabular}{p{3.2cm} P{0.88cm} P{0.88cm} P{0.88cm} P{0.88cm} }
			\toprule
			\multirow{2}{*}{\textbf{Method}} & \multicolumn{4}{c}{\textbf{LPIPS}($\uparrow$)} \\
			& 1&2 &3 &Avg.\\
			\hline 
			FCBoost-Net w/o $\mathcal{L}_{div}$ & 0.122 & 0.109 & 0.097 & 0.109 \\
			\hline
			FCBoost-Net (ours) & \textbf{0.521} & \textbf{0.499} & \textbf{0.454} & \textbf{0.491} \\
			\bottomrule
		\end{tabular}
	\end{table}
	
	\begin{figure}[t]
		\centering
		\includegraphics[width=0.5\textwidth]{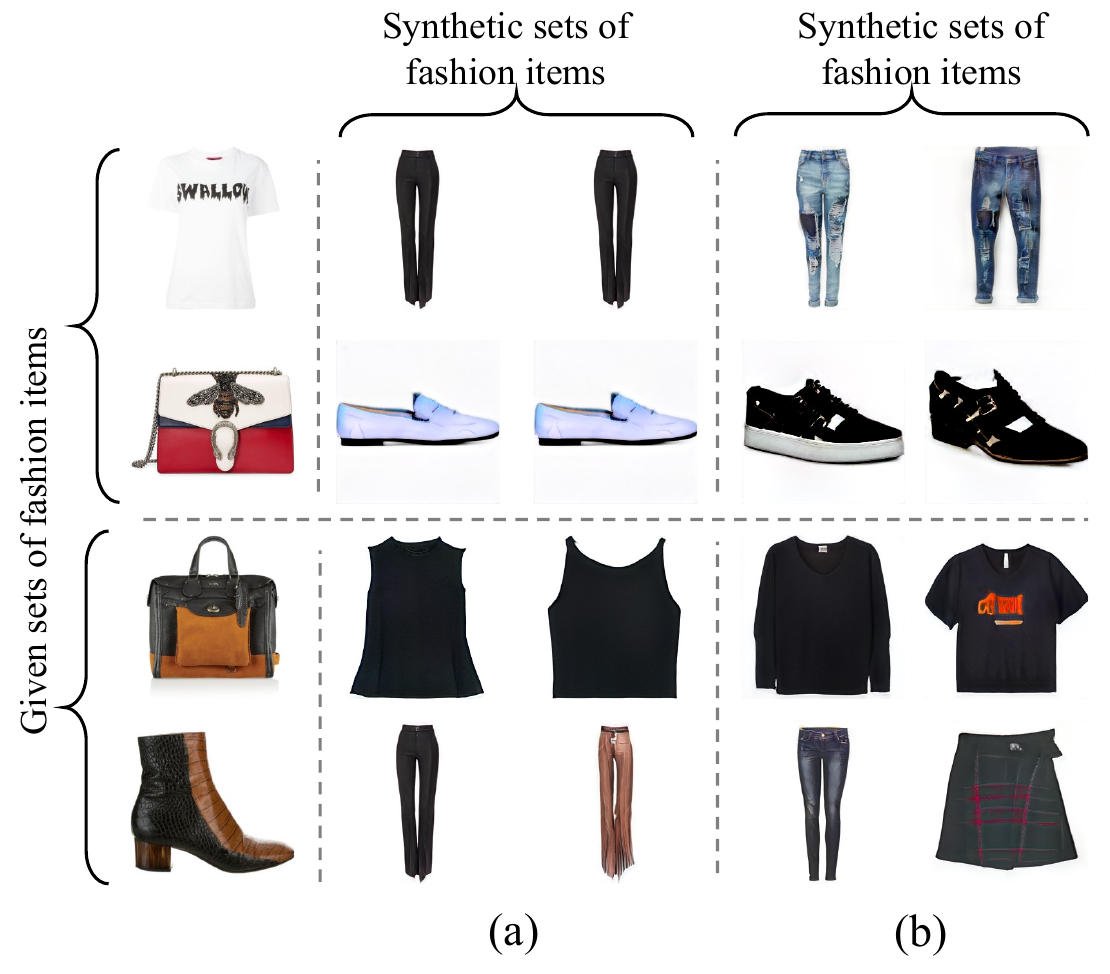}
		\caption{Visual comparison of the results for the effectiveness of the diversity loss function: (a) FCBoost-Net without diversity loss, and (b) FCBoost-Net with diversity loss (ours).}
		\label{fig:lpips_abla}
	\end{figure}
	\textbf{Effectiveness of the Diversity Loss.} The diversity loss function is used to improve the diversity of the sets of fashion items synthesized by FCBoost-Net. The efficacy of the diversity loss was assessed through a comprehensive analysis of its impact on the quantitative and qualitative performance, and a comparison of the results is shown in Table \ref{tab:lpips_abla}. The results indicate that the FCBoost-Net model with the diversity loss outperformed the model without the diversity loss by a significant margin (approximately 0.308 on average). To further support our quantitative findings, a qualitative evaluation of synthetic samples is presented in Fig. \ref{fig:lpips_abla}. The use of the diversity loss with FCBoost-Net can produce more diversified results than the version without it, especially in terms of the color and shape of the items. This indicates that the addition of this loss function significantly improves the diversity in terms of both quantitative and qualitative performance. 
	
	\begin{figure}[t]
		\centering
		\includegraphics[width=0.5\textwidth]{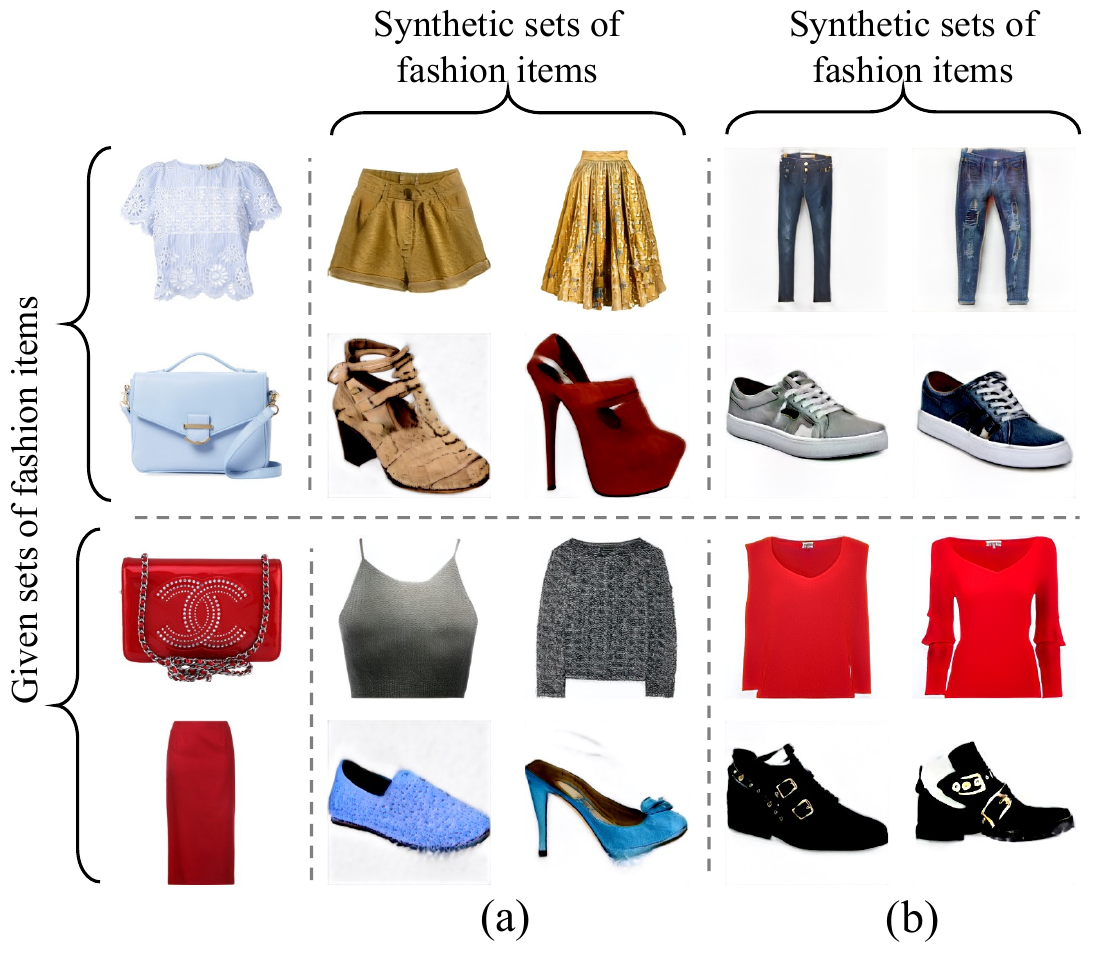}
		\caption{Visual comparison results for the effectiveness of fashion compatibility booster: (a) FCBoost-Net without fashion compatibility booster, and (b) FCBoost-Net with fashion compatibility booster (ours).}
		\label{fig:f2bt_abla}
	\end{figure}
	
	\begin{table}[t]
		\centering
		\small
		\caption{Comparison of the compatibility of the results of FCBoost-Net with other variants without the fashion compatibility booster module (F$^2$BT)}
		\label{tab:f2bt_abla}
		\begin{tabular}{p{3.7cm} P{0.76cm} P{0.76cm} P{0.76cm} P{0.76cm} }
			\toprule
			\multirow{2}{*}{\textbf{Method}} & \multicolumn{4}{c}{\textbf{F$^2$BT}($\uparrow$)} \\
			& 1&2 &3 &Avg.\\
			\hline 
			FCBoost-Net w/o $\mathcal{L}_{fcb}$ & 16.1\% & 19.5\% & 23.3\% & 19.6\% \\
			FCBoost-Net w/ $\mathcal{L}_{cmp}$ \cite{zhou2022coutfitgan} & 16.4\% & 21.9\% & 25.0\% & 21.1\% \\
			\hline
			FCBoost-Net (ours) & \textbf{33.1\%} & \textbf{30.6\%} & \textbf{29.3\%} & \textbf{31.0\%} \\
			\bottomrule
		\end{tabular}
	\end{table}
	
	\textbf{Effectiveness of the Fashion Compatibility Booster.} The fashion compatibility booster was designed to improve the compatibility of the current synthesized outfits in a progressive manner. The effectiveness of the module was validated from two perspectives. In the first experiment, FCBoost-Net was trained without the use of the fashion compatibility booster, and its performance was compared with the original FCBoost-Net based on the F$^2$BT metric. The results in Table \ref{tab:f2bt_abla} demonstrate that the fashion compatibility booster gave an improvement in F$^2$BT of 11.4\% on average, thus highlighting its importance in providing visually collocated supervision during the training of FCBoost-Net. We also generated some samples using FCBoost-Net with and without the fashion compatibility booster, as shown in Fig. \ref{fig:f2bt_abla}. It can be seen that with the proposed fashion compatibility booster, FCBoost-Net gave colors and textures that harmonized better with the given set of fashion items. In the second experiment, we compared our fashion compatibility booster with the fashion compatibility discriminator from COutfitGAN \cite{zhou2022coutfitgan}. A comparison of the results is presented in Table \ref{fig:f2bt_abla}, and it is clear that our proposed fashion compatibility booster has better supervision ability in comparison to that of COutfitGAN, with an improvement of approximately 10.0\% on average. The inclusion of the fashion compatibility discriminator from COutfitGAN in the FCBoost-Net architecture resulted in only a minimal improvement of around 1.5\% on average, compared to the version without fashion compatibility supervision. Our conjecture is that the fashion compatibility discriminator from COutfitGAN may have caused the creation of fashion items that were too similar, as it was designed to ensure that fashion items in an outfit have similar styles. This objective conflicts with our goal of generating collocated and diverse outfits.
	
	\begin{figure}[t]
		\centering
		\includegraphics[width=0.5\textwidth]{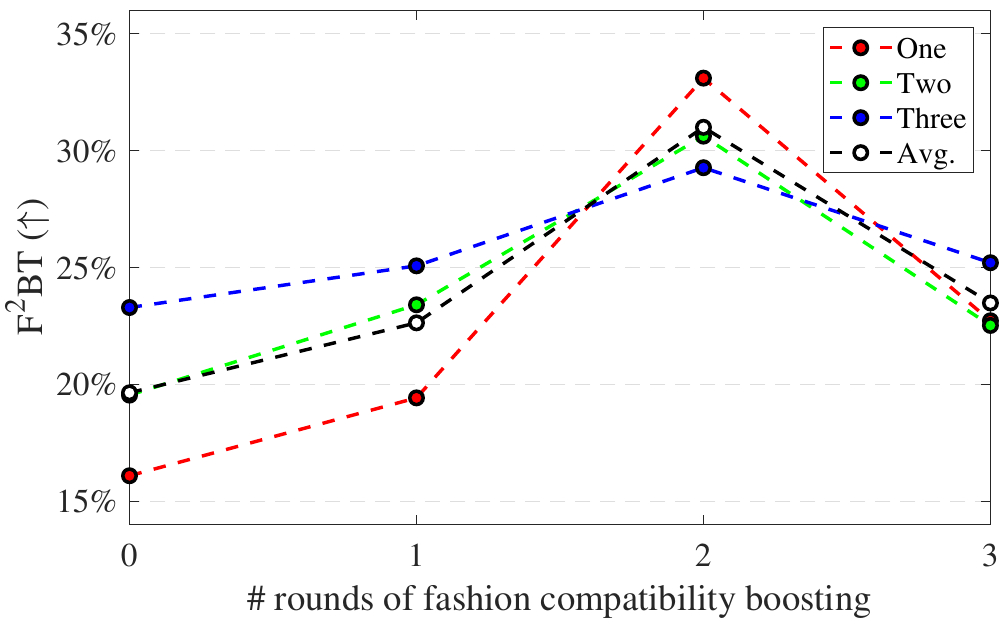}
		\caption{Fashion compatibility measurements for varying numbers of given fashion items and varying numbers of boosting rounds.}
		\label{fig:f2bt4rounds}
	\end{figure}

	\textbf{Effectiveness of Boosting.} To investigate the impact of the number of rounds of boosting, we present the results from FCBoost-Net with varying numbers of rounds from $\{0,1,2,3\}$.
	The results for the F$^2$BT metric are presented in Fig. \ref{fig:f2bt4rounds}. Here, the use of zero rounds means that we trained FCBoost-Net without the use of the fashion compatibility booster. From Fig. \ref{fig:f2bt4rounds}, we can see that an increase in the number of rounds within the range $[0,2]$ generally leads to an improvement in the performance of FCBoost-Net in terms of F$^2$BT. When the number of boosting rounds is greater than two, the performance of FCBoost-Net may become slightly worse. We ascribe this to the fact that the sets in the OutfitSet dataset consist of four categories of fashion items, and the number of given fashion items is at least one. The big number of boosting rounds may lead our model overfitted to the training set. In our experiments, setting the number of boosting rounds to two was sufficient to deliver satisfactory results.

	\section{Conclusion and Future Work}
	\label{sec:cnc}
	This paper has addressed a new research problem in the field of fashion synthesis, which involves the generation of collocated and diversified sets of fashion items based on a given set. To tackle this challenge, we proposed a novel framework called FCBoost-Net, which incorporates a pre-trained generative model with encoding networks to generate latent codes that globally control the generation of fashion items regardless of the spatial non-alignment in an outfit. To enhance the compatibility and diversity of the synthesized fashion sets, we also developed a boosting algorithm. Extensive experimental results demonstrate that our approach outperforms existing methods and achieves state-of-the-art performance in outfit generation. In the future, we plan to explore the use of guidance from other modalities to produce even more controllable and diverse images of fashion items.
	
	\textbf{Acknowledgments.} This work was supported in part by the National Natural Science Foundation of China under Grant no. 6197 2112, no. 61832004, and no. 62106211, the Guangdong Basic and Applied Basic Research Foundation under Grant no. 2021B1515020088, the Shenzhen Science and Technology Program under Grant no. JCYJ20210324131203009,  the Shenzhen Research Institute of Big Data Research Foundation under Grant no. T00120210002, the HITS Z-J\&A Joint Laboratory of Digital Design and Intelligent Fabrication under Grant no. HITSZ-J\&A-2021A01.
	\bibliographystyle{ACM-Reference-Format}
	\balance
	\bibliography{references}

	\appendix

\end{document}